\documentclass{article} % For LaTeX2e
\usepackage{iclr2024_conference,times}

\usepackage{amsmath,amsfonts,bm}

\def\eqref#1{equation~\ref{#1}}
\def\1{\bm{1}}

\DeclareMathAlphabet{\mathsfit}{\encodingdefault}{\sfdefault}{m}{sl}
\SetMathAlphabet{\mathsfit}{bold}{\encodingdefault}{\sfdefault}{bx}{n}

\DeclareMathOperator*{\argmax}{arg\,max}
\DeclareMathOperator*{\argmin}{arg\,min}

\usepackage{amsmath}
\usepackage{amsfonts}
\usepackage{amsthm}
\usepackage{bbm}
\usepackage{bm}
\usepackage{multirow}
\usepackage{booktabs}
\usepackage{graphicx}
\usepackage{caption}
\usepackage{subcaption}
\usepackage{float}
\usepackage[ruled]{algorithm2e}
\usepackage[colorlinks=true, allcolors=blue]{hyperref}
\usepackage{adjustbox}
\usepackage{wrapfig}
\usepackage{enumitem}

\newtheorem{theorem}{Theorem}[section]
\newtheorem{assumption}{Assumption}[section]

\newtheorem{lemma}[theorem]{Lemma}

\newcommand{\mcA}{\mathcal{A}} 
\newcommand{\mcB}{\mathcal{B}}

\newcommand{\bracnum}{N_{[]}}

\usepackage{hyperref}
\usepackage{url}

\title{A Generalization Theory of Cross-Modality Distillation with Contrastive Learning}
\author{Hangyu Lin$^{1}$, Chen Liu$^{1}$, Chengming Xu$^{2}$, Zhengqi Gao$^{3}$, Yanwei Fu$^{2}$, Yuan Yao$^{1}$\thanks{Correspond to: Yuan Yao $<$yuany@ust.hk$>$.} \\
$^{1}$The Hong Kong University of Science and Technology\\
\texttt{\{hlinbh,cliudh\}@connect.ust.hk, yuany@ust.hk} \\
$^{2}$Fudan University\\
\texttt{\{cmxu18, yanweifu\}@fudan.edu.cn} \\
$^{3}$Massachusetts Institute of Technology \\
\texttt{zhengqi@mit.edu} \\
}

\iclrfinalcopy

\begin{document}

\maketitle

\begin{abstract}
Cross-modality distillation arises as an important topic for data modalities containing limited knowledge such as depth maps and high-quality sketches. Such techniques are of great importance, especially for memory and privacy-restricted scenarios where labeled training data is generally unavailable. 
To solve the problem, existing label-free methods leverage a few pairwise unlabeled data to distill the knowledge by aligning features or statistics between the source and target modalities. 
For instance, one typically aims to minimize the L2 distance or contrastive loss between the learned features of pairs of samples in the source (e.g. image) and the target (e.g. sketch) modalities. However, most algorithms in this domain only focus on the experimental results but lack theoretical insight. To bridge the gap between the theory and practical method of cross-modality distillation,  we first formulate a general framework of cross-modality contrastive distillation (CMCD), built upon contrastive learning that leverages both positive and negative correspondence, towards a better distillation of generalizable features. Furthermore, we establish a thorough convergence analysis that reveals that the distance between source and target modalities significantly impacts the test error on downstream tasks within the target modality which is also validated by the empirical results. Extensive experimental results show that our algorithm outperforms existing algorithms consistently by a margin of 2-3\% across diverse modalities and tasks, covering modalities of image, sketch, depth map, and audio and tasks of recognition and segmentation. 
\end{abstract}

\section{Introduction}
% \begin{wrapfigure}{r}{0.45\textwidth}
%   \centering
%   \vspace{-10pt}
%   \includegraphics[width=0.45\textwidth]{figs/t-sne-3.pdf}
%   \caption{t-SNE embeddings of feature from direct SimCLR~\citep{chen2020simple} (top), CMKD~\citep{zhao2020knowledge}(bottom-left), and our CMCD (bottom-right).}
%   % \textcolor{red}{YW: try to visualize only3 or 4 classes in this figure for each method.}
%   \label{fig:t-sne}
%   \vspace{-10pt}
% \end{wrapfigure}
Cross-modality distillation is a significant topic in machine learning and deep learning, which distills the 'rich' knowledge in one modality to improve the modality of 'limited' knowledge~\citep{patel2015visual,gupta2016cross,thoker2019cross,ahmed2022cross}. However, most existing cross-modality methods~\citep{patel2015visual,saito2020universal,liang2020we,gupta2016cross,thoker2019cross,wang2022cross} require labeled data in the source modality, which may not be available in real scenarios due to memory or privacy constraints.
To address this challenge, some recent approaches~\cite {zhao2020knowledge,ahmed2022cross}  propose to distill knowledge without label information and align  features or statistics between the source and target modalities using a few paired data instead.
For instance, given images as the source modality and sketches as the target,
a common method is to minimize the L2 distance~\citep{gupta2016cross} or contrastive~\citep{radford2021clip,ahmed2022cross,wang2023one} loss between the learned features extracted from the images and sketches. Recently, contrastive learning methods have achieved many successful applications in cross-modality circumstances. CLIP~\citep{radford2021clip}, one of the most remarkable cross-modality models, achieves the alignment of image and caption representations in a shared feature space through the application of a contrastive loss which involves comparing pairs of images and captions and optimizing their alignment. In the model of One-Peace~\citep{wang2023one}, they try to explore the unified representation of multi-modalities by using the cross-modality aligning with contrastive and intra-modality denoising contrastive losses. However, these algorithms mainly focus on the experimental results but lack the theoretical analysis of the methods. 

To fill the gap, we first formulate a versatile framework called generalizable Cross-Modality Contrastive Distillation (CMCD) based on contrastive learning in this paper. The CMCD framework tries to investigate both the positive and negative relationships in a contrastive distillation way that enables the efficient transfer of generalizable features from the source modality to the target modality. 
Concretely, we introduce two types of loss functions based on contrastive learning for conducting cross-modality contrastive distillation. One of them is designed upon knowledge distillation techniques and the other is inspired by multi-modality pretraining works such as CLIP\citep{radford2021clip}. 
One well-known theoretical result in transfer learning~\citep{ben2010theory} states that when applying domain adaptation methods, the test error of the target modality can be bounded by the test error of the source modality and the $\mathcal{H}\Delta\mathcal{H}$ divergence between the source and target modalities. However, this result is not directly applicable to contrastive learning-based algorithms. 
Moreover, existing theoretical research on contrastive learning~\citep{tosh2021contrastive,saunshi2022understanding,ge2023provable} primarily focuses on single modality scenarios. Therefore, to analyze our algorithm, we develop novel theories that merge cross-modality learning and contrastive learning.
Through our analysis, we give the generalization bound of each step in our algorithm and introduce a final theorem that illustrates the final test error of the downstream task in the target modality will be bounded by the total variation distance between the source and target modalities. In essence, the theorem reveals that as the total variation distance between the modalities decreases, our algorithm has a higher likelihood of obtaining generalizable features in the target modality. This theoretical insight is further validated by our experimental results. Besides the theoretical analysis, we demonstrate that our algorithms can outperform other cross-modality transfer algorithms across various modalities (e.g., images, sketches, depth maps, videos, and audio) and tasks such as recognition and segmentation. 

In summary, our contributions are as follows:
% We state our contributions as follows:

\begin{itemize}
    \item  A generalizable cross-modality contrastive distillation (CMCD) framework is formulated to utilize both positive and negative relationships in paired data and effectively distill generalizable representation from a source modality to a specific target modality.
    \item We perform a theoretical analysis of our algorithm, elucidating the algorithm's convergence bounds. Our findings underscore a direct correlation between the algorithm's ultimate performance and the total variation distance between the source and target modalities that is further validated by our empirical results.
    \item We conducted extensive experiments that provide compelling evidence for the efficacy of our algorithm across diverse modalities and a range of downstream tasks.
    
    % A Rigorous analysis of our algorithm is provided to show the convergence bound of our algorithm. The final result shows that the final performance of the algorithm is related to the total variation distance between source and target modalities which is also demonstrated by the empirical results.
\end{itemize}
\section{Related Works}

\noindent \textbf{Cross-modality transfer/distillation}.
Cross-modality distillation, which can be viewed as a form of transfer learning, has been a subject of study for a considerable period~\citep{gupta2016cross,zhao2020knowledge,xue2021multimodal,xue2022modality}. In the realm of deep learning, \cite{gupta2016cross} propose a distillation method that relies on feature alignment to transfer supervision from RGB images to depth images. Building upon this work, CMKG~\citep{zhao2020knowledge} encoded teacher network knowledge into priors through meta-learning. SOCKET~\citep{ahmed2022cross} proposes to minimize the distance between the statistics of features instead of the original feature activation maps. Rather than focusing on an unimodal student, MKE~\citep{xue2021multimodal} leveraged multimodal pairs to enhance the student network. \cite{wang2023one} introduced a generic pretraining framework that incorporates knowledge from multiple modalities using cross-modality contrastive learning. However, their method necessitates a substantial amount of paired data. 
Another line of research explores dataset distillation~\citep{lei2023comprehensive}, which aims to select or generate the most informative samples for feature embedding in the target modality. In our research, we aim at transferring the knowledge from a source modality to a target modality which is both unsupervised and uses a small number of paired data to distill the knowledge.

\noindent \textbf{Contrastive learning}.
Contrastive learning has emerged as a popular and effective technique in self-supervised and unsupervised tasks~\citep{oord2018representation,wu2018unsupervised,chen2020simple,he2020momentum}. At its core, contrastive learning trains models to capture discriminative features among data instances, typically achieved through data augmentation~\citep{chen2020simple} and the InfoNCE loss~\citep{oord2018representation}. Methods such as SimCLR/SimCLR v2~\citep{chen2020simple,chen2020big} propose simple frameworks with various data augmentations for contrastive learning. Momentum Contrast~\citep{he2020momentum} introduced a momentum encoder to enhance negative pairs' features from a memory bank. Additionally, BYOL~\citep{grill2020bootstrap} maintains online and target networks, updating the target network with a slow-moving average of the online network.
% Most existing methods and studies on contrastive learning focus on conducting self-supervised learning or pre-training on large datasets within a single modality, such as images or natural language.
More recently, foundational models like CLIP~\citep{radford2021clip} have connected images and language through contrastive learning. However, contrastive learning approaches are primarily designed for self-supervised pre-training on large-scale datasets. In this paper, we explore how contrastive learning can serve as a powerful method for transferring knowledge from a major modality, such as images, to a minor modality, such as high-quality sketches.

% \noindent \textbf{Theoretical Results of Contrastive Learning and Transfer Learning}
\noindent \textbf{Theoretical results}. We discuss the  theoretical results of Contrastive Learning and Transfer Learning.
Compared to the experimental research, there is a limited theoretical analysis of cross-modality knowledge distillation. MFH~\citep{xue2022modality} proposes a hypothesis that emphasizes the importance of overlapped information between two domains for successful distillation, supported by a simple proof in the linear case. To contextualize our research, we also review theoretical results from the fields of contrastive learning and transfer learning. \cite{wen2021toward} prove that contrastive learning with ReLU activation can learn desirable sparse features when appropriate augmentations are used. In \cite{saunshi2022understanding}, they elaborate on the importance of inductive biases in the analysis of contrastive learning. The most related analysis to our research is \cite{ge2023provable}, they give a complexity-based convergence bound of contrastive learning which indicates the benefits of the pretraining but is not concerned about the cross-modality contrastive case. For transfer learning, a basic theoretical result from \cite{ben2010theory} states that target error can be bounded by the source error and the $\mathcal{H}\Delta\mathcal{H}$ divergence between source and target distributions under a supervised domain adaptation setting. \cite{tripuraneni2020theory} formulate the target error by using the task-representation difference which can be used to measure the task diversity. 
% In this paper, we model the error bound in the target modality by using the total variation between the source and target distributions constructed by contrastive learning.

\section{Methodology}
% \subsection{Problem Setup}
\noindent\textbf{Notations.} We define the source modality and target modality as $\mathcal{A}$ and $\mathcal{B}$. We denote by $x_{\mcA} \in \mathcal{X}_{\mcA},  x_{\mcB} \in \mathcal{X}_{\mcB}$ the input data of two modalities respectively. In our setting, we do not require the supervised labels for both the source modality $\mcA$ and target modality $\mcB$ to perform cross-modality distillation. Subsequently, when performing a downstream task in modality $\mcB$, a few labels $y_{\mcB} \in \mathcal{Y}_{\mcB}$ are needed for fine-tuning the model. Our goal is to get an efficient model of the downstream task in target modality $\mcB$, i.e., a model can predict $y_{\mcB}$ from $x_{\mcB}$. We assume that $y$ is connected to $x$ through a latent variable/feature $z$ which means that given $z$ the value of $y$ is independent of $x$. In order to construct the theoretical analysis for our algorithm, we follow the setting of \cite{ge2023provable} and introduce the side information $s \in \mathcal{S}$ which can be accessed from $x \in \mathcal{X}$. For example, in contrastive learning, given $(x, x^{\prime}) \in \mathcal{X}^2$, $s:= \mathbbm{1}(x = x^{\prime})$ where the side information indicates whether the pair should be considered as positive or negative. To measure the distance between two distributions, we use total variation distance $d_{TV}(\mathbb{P}, \mathbb{Q}) = \frac{1}{2}\int |p(x)-q(x)| dx$.

We use two kinds of models to construct distributions of $(x, z, y)$, the latent variable model $\phi$ modeling the relationship between $(x, z)$ and the prediction model $\psi$ modeling the relationship between $(z, y)$. Concretely, we assume that there exist oracle models $\phi_{\mcA}^{*}, \phi_{\mcB}^{*}, \psi_{\mcB}^{*}$ for both source and target modalities $\mathcal{A}, \mathcal{B}$, indicating $z_{\mcA} =\phi_{\mcA}^{*} (x_{\mcA} )$, $ z_{\mcB} =\phi_{\mcB}^{*} (x_{\mcB} )$, $y_{\mcB} = \psi_{\mcB}^{*}(z_{\mcB}) $. 

\subsection{Framework of Generalizable Cross-Modality Contrastive Distillation}
\paragraph{Step 1.} Firstly, given the source modality $\mathcal{A}$ with massive unlabeled data $S_{\mcA} = {\{x_i^{\mcA}\}_{n_{\mcA}}}$, we can use typical contrastive learning such as SimCLR to learn the latent feature representation $\hat{\phi}_{\mcA}$. Specifically, we use the InfoNCE~\citep{wu2018unsupervised} loss to train the model $\phi_{\mcA}$:
\begin{align}
    \mathcal{L}_{\text{InfoNCE}} = - \sum_{i,j} \log \frac{\exp(z_i^{\mathcal{A}} \cdot z_j^{\mathcal{A}}/\tau)}{\sum_{t} \exp(z_t^{\mathcal{A}} \cdot z_j^{\mathcal{A}}/\tau)}
\end{align}
where $\tau$ is the temperature hyper-parameter, and $z^{\mcA}_{i} = \phi_{\mcA}(x_i^{\mcA})$ is a projected feature by the model $\phi_{\mcA}$. Numerous studies~\citep{chen2020simple,he2020momentum,grill2020bootstrap} have demonstrated that self-supervised models can serve as effective feature extractors for various downstream tasks.
\paragraph{Step 2.} Secondly, we leverage the pair data of source and target modalities $S_{\mcA\mcB} = \{(x_i^{\mcA}, x_i^{\mcB})\}_m$ to distill the information from the source modality to the target modality. In this cross-modality distillation step, we propose two types of cross-modality losses. The first one is based on knowledge distillation and is referred to as the cross-modality distillation (CMD) loss:

\begin{align}
    \mathcal{L}_{\text{CMD}} = - \sum_{i,j} \frac{\exp(z_i^{\mathcal{A}} \cdot z_j^{\mathcal{A}}/\tau)}{\sum_{t} \exp(z_t^{\mathcal{A}} \cdot z_j^{\mathcal{A}}/\tau)} \log \frac{\exp(z_i^{\mathcal{B}} \cdot z_j^{\mathcal{B}}/\tau)}{\sum_{t} \exp(z_t^{\mathcal{B}} \cdot z_j^{\mathcal{B}}/\tau)}
\end{align}

where $z^{\mcA}_{i} = \hat{\phi}_{\mcA}(x_i^{\mcA})$ and $ z^{\mcB}_{i} = \phi_{\mcB}(x_i^{\mcB})$ represent the learned feature in source modality and the project feature from target modality to be optimized. This distillation-type loss is proposed for self-supervised distillation first~\citep{fang2021seed} which aims at distilling the information from a large model (e.g., ResNet101) to a small model (e.g., ResNet18) without any supervision. The second loss is inspired by CLIP~\citep{radford2021clip} and is called  cross-modality contrastive (CMC) loss:
\vspace{-3pt}
\begin{align}
    \mathcal{L}_{\text{CMC}} = - \sum_{i} \left(\log \frac{\exp(z_i^{\mathcal{A}} \cdot z_i^{\mathcal{B}}/\tau)}{\sum_{t} \exp(z_t^{\mathcal{A}} \cdot z_i^{\mathcal{B}}/\tau)} +  \log \frac{\exp(z_i^{\mathcal{B}} \cdot z_i^{\mathcal{A}}/\tau)}{\sum_{t} \exp(z_t^{\mathcal{B}} \cdot z_i^{\mathcal{A}}/\tau)}\right)
\end{align}
This loss was originally used for multi-modality pretraining which needs a lot of paired data, while in our algorithm we utilize it to transfer latent features from the source modality to the target modality which needs much less paired data than pretraining. As will be demonstrated, both CMD and CMC losses work well for cross-modality knowledge distillation theoretically and experimentally.
\paragraph{Step 3.} After distillation, we can use the learned feature representation $\hat{\phi}_{\mcB}$ in the target modality to solve downstream tasks (e.g., classification, semantic segmentation) with some simple fine-tuning, i.e., training a one-layer classifier based on the features. For instance, we can utilize the model $\hat{\phi}_{\mcB}$ and a small number of labels $y_i^{\mcB}$ to train an MLP for a classification task. We formulate this step using cross entropy loss in Step 3 in Algorithm~\ref{alg}. However, this task can be replaced by any other downstream tasks such as semantic segmentation or detection with different labels and loss functions.
The overall algorithm flow is summarized in Algorithm~\ref{alg} below. 

\begin{algorithm}
  \SetAlgoLined
  \KwData{$S_{\mathcal{A}} = {\{x_i^{\mcA}\}_{n_{\mcA}}}, S_{\mcB} = {\{(x_i^{\mcB}, y_i^{\mcB})\}_{n_{\mcB}}}, S_{{\mcA}{\mcB}} = \{(x_i^{\mcA}, x_i^{\mcB})\}_m$}
  \KwResult{$\hat{\phi}_{\mcB},  \hat{\psi}_{\mcB}, \hat{\phi}_{\mcA} $}

  \textbf{Step 1}: Contrastive learning of source modality $\mathcal{A}$
    \begin{equation}
        \hat{\phi}_{\mcA} = \argmin_{\phi \in \Phi_{\mcA}} \sum_{i,j=1}^{n_{\mcA}} \mathcal{L}_{\text{InfoNCE}}(\phi(x_{i}), \phi(x_{j}), s_{ij})
        \label{eq:step1_1}
    \end{equation}
        
  \textbf{Step 2}: Distillation: contrastive distillation of $\mathcal{A}, \mathcal{B}$ to an error of $\epsilon_{\mcA\mcB}$
    \begin{equation}
    \hat{\phi}_{\mcB} = \argmin_{\phi \in \Phi_{\mcB}} \sum_{i,j=1}^{n} \mathcal{L}_{\text{CM}}( \phi(x_{i}),  \phi(x_{j}), { \hat{\phi}_{\mcA}}(x_{i}), {\hat{\phi}_{\mcA}}(x_{j}))\, 
    % , \quad (\mathcal{L}_{\text{CM}}=\mathcal{L}_{\text{CMD}} \text{ or } \mathcal{L}_{\text{CMC}})
    \label{eq:step2}
    \end{equation}
    
  \textbf{Step 3}: fine-tune on the target modality $\mathcal{B}$ to an error of $\epsilon_{\mcB}$
    \begin{equation}
    \hat{\psi}_{\mcB} = \argmin_{\psi \in \Psi_{\mcB}} \sum_{i=1}^{n_{\mcB}} \mathcal{L}_{\text{CE}}({\psi \circ  \hat{\phi}_{\mcB} }(x_i), y_i)
    \label{eq:step3}
    \end{equation}
  \caption{Generalizable Cross-Modality Contrastive Distillation}
  \label{alg}
\end{algorithm}

\subsection{Theoretical Analysis}
In this section, we prove that the test error of the downstream task in target modality $\mcB$ can be bounded in probability by the total variation of the latent feature distribution between source and target modality, i.e.,$d_{TV}(\mathbb{P}_{{\phi}_{\mcB}^{*}}, \mathbb{P}_{{\phi}_{\mcA}^{*}})$ and the Rademacher complexities related to $\Phi_{\mcB}$ and $\Psi_{\mcB}$, respectively. We mainly analyze the CMD loss in the main text and the discussion for CMC loss can be found in Appendix~\ref{app:cmc}. To begin with, we introduce an assumption that builds a relationship between the contrastive loss and the downstream loss.

\begin{assumption}
    \label{ass1}
    ($\kappa^{-1}$-informative condition.) We assume that the model class $\Phi$ is $\kappa^{-1}$-informative with respect to the true models $\phi^{*}, \psi^{*}$ if for any $\phi \in \Phi$, and $x \in \mathcal{X}$,  such that
    \begin{equation}
        \mathcal{L}_{\text{CE}}({ {\psi}^{*} \circ {\phi}}(x), y) \le \kappa \mathbb{E}_{x^{\prime}}[\mathcal{L}_{\text{CMD}}({\phi}, {\phi}^{*},(x, x^{\prime}), s)]
        % d_{TV}(\mathbb{P}_{T_1 \circ \phi }(x, z), \mathbb{P}_{\phi^{*}}(x, z)) \le \kappa \cdot d_{TV}(\mathbb{P}_{\phi}(x, s), \mathbb{P}_{\phi^{*}}(x, s))
    \end{equation}
    where $\mathcal{L}_{\text{CM}}({\phi}, {\phi}^{*},(x, x^{\prime}), s):=\mathcal{L}_{\text{CM}}({\phi}(x), {\phi}(x^{\prime}), {\phi}^{*}(x), {\phi}^{*}(x^{\prime}))$ for notation simplicity.
\end{assumption}

It is introduced in \cite{ge2023provable} to guarantee the feature extraction model $\phi$ and the side information $s$ contains a certain level of information that can reveal the relationship between $x$ and $z$. In other words, this assumption implies that the model obtained with contrastive learning performs reasonably well on downstream tasks. Here we assume that it holds true for both source and target modalities. 

% \subsection{Bound of Contrastive Distillation Step}

In order to derive the whole generalization bound of our algorithm, we start with the bound of the contrastive learning step, i.e., Step 1 of Algorithm~\ref{alg}.
\begin{lemma}
    \label{lem:con_tv}
    Let $\hat{\phi}_{\mcA}$ the minimizer of \eqref{eq:step1_1}. Then, with probability at least $1 - \delta$, we have, 
    \begin{equation}
        d_{TV}(\mathbb{P}_{\hat{\phi}_{\mcA}}(\bm{x}, s), \mathbb{P}_{{\phi}^{*}_{\mcA}}(\bm{x}, s)) \le 3 \sqrt{\frac{1}{{n_{\mcA}}^2} \log \frac{\bracnum(P_{\mathcal{X}_{\mcA}  \times \mathcal{S}}(\Phi_{\mcA}), \frac{1}{{n_{\mcA}}^2})}{\delta}}
    \end{equation}
    where $\mathcal{P}_{\mathcal{X}_{\mcA}  \times \mathcal{S}}(\Phi_{\mcA}) = \{\mathbb{P}_{\phi_{\mcA}}(\bm{x}, s)|\phi_{\mcA} \in \Phi_{\mcA}\}$, $\bm{x} = (x_i, x_j)$, $s$ indicates whether $x_i, x_j$ is the paired data,  and $N_{[]}(\cdot, \cdot)$ denotes the bracket number.
    Here the density function of the distribution $\mathbb{P}_{\phi}(\bm{x}, s)$ is defined by,
\begin{align}
    p_{\phi}(\bm{x}, s) = \frac{\exp(z_i \cdot z_j/\tau)}{\sum_{t} \exp(z_t \cdot z_j/\tau)}; z_i = \phi(x_i)
\end{align}
\end{lemma}

which can be seen as a Gibbs distribution of modeling the paired data in a contrastive learning view. This lemma states that with contrastive learning the total variation distance between the best feature representation $\phi_{\mcA}^{*}$ and the learned feature representation $\hat{\phi}_{\mcA}$ can be bounded by the bracket number of the possible distribution space $\mathcal{P}_{\mathcal{X}_{\mcA}\times \mathcal{S}}(\Phi_{\mcA})$. The proof of this lemma is based on a reformulation of the contrastive learning task into a maximum likelihood estimation task. Detailed proof of this lemma can be found in Appendix~\ref{sec:proof_lem_tv}

Now we bound the test error of the CMD loss learned from the contrastive distillation step, i.e., Step 2 of Algorithm~\ref{alg}.
\begin{theorem}
    \label{thm:dist}
    Let $\hat{\phi}_{\mcB}$ the minimizer of \eqref{eq:step2}. Assume that $\sup_{\phi_{\mcB} \in \Phi_{\mcB}, \bm{x}_{ij}} \langle {\phi_{\mcB}}(\bm{x}_i), {\phi_{\mcB}}(\bm{x}_j)\rangle  \le B$. Then given $\hat{\phi}_{\mcA}$, with probability at least $1 - \delta$, we have 
    \begin{equation}
        \mathbb{E}[\mathcal{L}(\hat{\phi}_{\mcA}, \hat{\phi}_{\mcB}, \bm{x}, s)] - \frac{1}{m^2}\sum_{i,j=1}^{m}\mathcal{L}(\hat{\phi}_{\mcA}, \hat{\phi}_{\mcB}, \bm{x}_{ij},s_{ij}) \le 2 R_{m^2}(\mathcal{L} \circ {\Phi_{\mcB}}) + B\sqrt{\frac{2\ln (1/ \delta)}{m^2}}
    \end{equation}
    where $\mathcal{L}(\hat{\phi}_{\mcB}, \hat{\phi}_{\mcA}, \bm{x}_{ij}, s_{ij}):= \mathcal{L}_{\text{CMD}} (\phi_{\mcB}(x_{i}), \phi_{\mcB}(x_{j}), \phi_{\mcA}(x_{i}), \phi_{\mcA}(x_{j}))$ for notation simplicity. 
\end{theorem}
This theorem describes a similar result of the oracle inequality of ERM where the difference between the test and empirical loss can be bounded by the Rademacher complexity $R_{m^2}(\mathcal{L} \circ {\Phi_{\mcB}})$ and an $\mathcal{O}(m^{-1})$ term.

\textit{Proof Sketch}: This bound comes from Mcdiarmid's inequality~\citep{zhang_2023} and the definition of the Rademacher complexity. By constructing the function $f(\bm{x}_{11}, \dots, \bm{x}_{mm}) =\sup_{\phi_{\mcB} \in \Phi_{\mcB}} [\mathbb{E}[\mathcal{L}(\hat{\phi}_{\mcA}, \hat{\phi}_{\mcB}, \bm{x}, s)] - \frac{1}{m^2}\sum_{i,j=1}^{m}\mathcal{L} (\hat{\phi}_{\mcA}, \hat{\phi}_{\mcB}, \bm{x}_{ij}, s_{ij})]$, we can prove that $f$ satisfy the bounded difference property and apply Mcdiarmid's inequality on $f$ to prove the final results. Detailed proof can be found in the Appendix~\ref{sec:proof_con_dist}.

% \subsection{Bound of Whole Algorithm}
Finally, we prove that the test error of the downstream task in target modality $\mcB$ can be bounded in probability by the total variation of the latent feature distribution between source and target modality, i.e.,$d_{TV}(\mathbb{P}_{{\phi}_{\mcB}^{*}}, \mathbb{P}_{{\phi}_{\mcA}^{*}})$ and the Rademacher complexities related to $\Phi_{\mcB}$, $\Psi_{\mcB}$ respectively. 

\begin{theorem}
    \label{thm:whole_alg}
    Let $\hat{\phi}_{\mcB}$ and $\hat{\psi}_{\mcB}$ be the outputs of Algorithm \ref{alg}. Suppose the loss function $\mathcal{L}$ is $L$-bounded, $\sup_{\phi_{\mcB} \in \Phi_{\mcB}, \bm{x}_{ij}} \langle {\phi_{\mcB}}({x}_i), {\phi_{\mcB}}({x}_j)\rangle  \le B$ and the model follows $\kappa^{-1}$-informative. Then with probability at least $1 - \delta$, the test risk of $\hat{\phi}_{\mcB}$ and $\hat{\psi}_{\mcB}$ is bounded by
    % \textcolor{red}{Need to fix the bound.}
    \begin{align}
        \mathbb{E}[\mathcal{L}({\hat{\psi}_{\mcB}} \circ \hat{\phi}_{\mcB}&(x), y)] \le \kappa B \cdot d_{TV}(\mathbb{P}_{{\phi}_{\mcB}^{*}}, \mathbb{P}_{{\phi}_{\mcA}^{*}}) + \kappa\epsilon_{\mcA\mcB} + \epsilon_{\mcB } \label{eq:uni_final_1}\\
        &\quad  + 2 \kappa R_{m^2}(\mathcal{L}_{\text{CMD}} \circ {\Phi_{\mcB}})   + 2 R_{n_{\mcB}}(\mathcal{L} \circ \Psi_{\mcB} \circ {\hat{\phi}_{\mcB}}) \label{eq:uni_final_2}\\ 
        &\quad + 3\kappa B \cdot  \sqrt{\frac{1}{{n_{\mcA}}^2} \ln \frac{4\bracnum(\mathcal{P}_{\mathcal{X}_{\mcA}  \times \mathcal{S}}(\Phi_{\mcA}), \frac{1}{{n_{\mcA}}^2})}{\delta}} + \kappa L\sqrt{\frac{2\ln (4 / \delta)}{m^2}} + 2L\sqrt{\frac{2\ln (4 / \delta)}{n_{\mcB}}} \label{eq:uni_final_3}
    \end{align}
\end{theorem}
Detailed proof of our the Theorem~\ref{thm:whole_alg} in Appendix~\ref{sec:proof_whole}.  Basically, there are three parts on the right side of the oracle inequality. The first part \eqref{eq:uni_final_1} is the total variation between the distributions of true $\phi_{\mcB}^{*}$ and $\phi_{\mcA}^{*}$ which represents the common information between modalities $\mathcal{B}, \mathcal{A}$. The error terms $\epsilon_{\mcA\mcB}, \epsilon_{\mcB}$ are the hyperparameters to control the empirical losses. The second term\eqref{eq:uni_final_2} measures the model complexity of the latent variable model $\Phi_{\mcB}$ and the prediction model ${\Psi_{\mcB}}$ respectively. Recall that direct ERM of the downstream task in the modality $\mathcal{B}$ is bounded by the complexity of the composition of the latent variable model and the prediction model $\Psi_{\mcB} \circ \Phi_{\mcB}$ which is much larger than the sum. When $m$ and $n_{\mcA}$ are much larger than $n_{\mcB}$ which is supposed in our task, the third term \eqref{eq:uni_final_3} demonstrates the convergence rate is close to the term in original ERM. Combine all the terms together, we can find that if the total variation between the distributions of true $\phi_{\mcB}^{*}$ and $\phi_{\mcA}^{*}$ is small, the final generalization bound is not worse or even better than  the bound of the supervised learning. It indicates that if source and target modalities have more common information or patterns, the algorithm will have a higher probability of distilling more information from the source modality to the downstream task in the target modality. We also validate this observation in experimental results.

\section{Experiments and Discussion}
\noindent \textbf{Experiments Setup}.
To demonstrate the effectiveness of our algorithm, we conduct extensive experiments on various cross-modality tasks. 

\textit{Image-sketch}: Here the image is the source modality and the sketch serves as the target modality. Specifically, we use the ImageNet~\citep{deng2009imagenet} as the source dataset $S_{\mcA}$ for contrastive learning and Sketchy~\citep{sangkloy2016sketchy} as the paired data $S_{\mcA\mcB}$. The downstream recognition task is performed on both Sketchy and TUBerlin datasets to evaluate the generalization performance of the algorithms. 

\textit{Video-audio}: Here the video and audio clips become the source and target modalities respectively. We still use ImageNet for contrastive learning on each frame of the video and then distill on a 4.6k subset of VGGSound~\citep{chen2020vggsound}. Random sampled 10k of the rest of the sound data works as the fine-tuning dataset of event recognition in the audio modality. 

\textit{Image-depth map}: Here the image and depth map are source and target modalities. ImageNet is used as the source dataset for contrastive learning. We then distill on the NYU-Depth V2~\citep{silberman2012nyuv2}. Segmentation is conducted on the depth map only as the downstream task.

We mainly use ResNet~\citep{he2016deep} models to learn feature extractors. For instance, we may use ResNet50 for ImageNet self-supervised pre-training and distill it to ResNet18/Resnet50 of the target modality. The data augmentation method in SimCLR is used in our contrastive learning and cross-modality distillation. For all training, we use Adam~\citep{kingma2014adam} optimizer with different learning rates. Commonly, a multistep exponential decay scheduler will be used to adjust the learning rate during training. 
Detailed settings on the datasets and models can be found in Appendix~\ref{app:set_exp}.

\noindent \textbf{Baselines}.
% Models setting/Optimizers/Appendix.
In our experiments, we compare our algorithm with the following baselines:

(i). \textit{Sup FT}: Since the downstream task is supervised, we can take the direct training as the baseline. For instance, for the recognition task, we will use the same backbone as our algorithm, and directly minimize the cross-entropy loss.
(ii). \textit{SSL}: We also deploy self-supervised learning methods such as SimCLR~\citep{chen2020simple} straightly in the target modality. After the SSL, we may use linear evaluation (LM) or fine-tuning (FT) on the supervised data to get the final performance. 
% Actually, our algorithm is mainly designed for unsupervised cross-modality distillation it is reasonable to compare the self-supervised method. 
(iii). \textit{PreSSL}: In addition to the SSL method, we also test start from the SSL pre-trained model in source modality. In the Image-Sketch task, we can use the SSL pre-trained model of ImageNet as the initial model for SSL of sketch data.
(iv). \textit{CMST}: Cross Modality Supervision Transfer~\citep{gupta2016cross} is a method that restricts the distance between the mid-level semantic representation of source and target modalities. Specifically, we use the L2 norm to compute the distance between feature extractors of source and target modalities. 
(v). \textit{CMKD}: In \cite{zhao2020knowledge}, they propose to not only add L2 distance on activation maps but also attention maps between the modalities. We adopt the same settings in their work to implement and evaluate this method.
(vi). \textit{SOCKET~\citep{ahmed2022cross}}: This method does not directly add L2 norm loss on the activation maps but the statistics of the feature maps, e.g., mean and variance.

\subsection{Main Results}
\begin{table} 
    \centering
    \begin{tabular}{c|cc|c|cc}
    \toprule 
        \multirow{2}{*}{Tasks} & \multicolumn{2}{c|}{image-sketch} & video-audio & {image-depth map} \\ \cmidrule{2-5}
         & Sketchy & TUBerlin & VGGSound & NYU-Dpeth V2  \\ \midrule
        SSL + LE & 64.31 & 54.64 & 18.48 & 13.88  \\ 
        PreSSL + LE & 67.90 & 60.56 & 19.31 & 17.37 \\
        CMST  + LE & 68.20 & 62.54 & 21.12 & 17.33 \\ 
        CMKD + LE & 70.97 & 64.46 & 26.38 &  17.61 \\  
        SOCKET + LE & 71.33 & 64.22 & 25.87 &  17.13 \\ \hline
        CMD + LE (Ours) & 72.61 & 65.70 & \textbf{28.30} & 17.36 \\
        CMC + LE (Ours)& \textbf{73.24} & \textbf{68.72} & 28.27 & \textbf{18.35} \\ \midrule
        Sup FT & 83.90 & 74.48 & 32.67 & 23.73 \\ 
        SSL + FT & 83.01 & 74.80 & 32.10 & 18.22 \\ 
        PreSSL + FT & 83.75 &75.50  & 32.37 & 26.06 \\
        CMST + FT & 83.32 & 75.36  & 32.11 & 26.16 \\
        CMKD + FT & 84.87 & 75.84 & 34.42 & 26.60 \\ 
        SOCKET + FT & 84.93 & 75.48 & 34.15 & 26.33 \\  \hline
        CMD + FT (Ours) & 85.63 & \textbf{77.86} & 35.13 & 27.20 \\ 
        CMC + FT (Ours) & \textbf{87.54} & {77.44} & \textbf{35.37} & \textbf{27.93} \\ 
    \bottomrule
    \end{tabular}
    \caption{Main results of our method and other baselines on different cross-modality tasks. All the tasks use ResNet50 as the teacher network and ResNet50 as the student network to solve the downstream tasks. We use top-1 accuracy(\%) for recognition tasks and mean IoU (\%) for the segmentation task.
     \textbf{LE} means a linear evaluation on only the final classification layer; \textbf{FT} means fine-tuning the whole network.}
    \label{tab:main}
\end{table}
\textbf{Effectiveness of our algorithm}.As depicted in Table~\ref{tab:main}, our algorithm surpasses other baselines across various tasks. Specifically, when fine-tuning a one-layer MLP on image-sketch modalities, our method utilizing CMD/CMC loss achieves top-1 accuracies of 72.61\%/73.24\% on Sketchy, outperforming the best baseline by a margin of 2\%. Even when fine-tuning the entire network, our method with CMC loss maintains the best performance, showcasing the effectiveness of our cross-modality distillation approach. The results obtained on TUBerlin further demonstrate the versatility of the learned features, with our approach achieving the highest accuracies of 68.72\%/77.86\% through linear evaluation or full fine-tuning.
% \vspace{-4pt}

In addition to image-sketch modalities, our method proves effective in the context of video-audio tasks. Notably, even when utilizing aggregated frame features, our method successfully distills meaningful features from the video/image modality to the audio modality. In the case of linear evaluation with audio data, our method exhibits a 10\% advantage over the original SSL method and a 2\% margin compared to the best baseline.
% \vspace{-4pt}

Moreover, the results obtained in image-depth segmentation highlight the efficacy of our algorithm. Although our method with linear evaluation or full fine-tuning remains the best approach, the improvement over other baselines is not substantial. This can be attributed to the fact that contrastive learning is primarily designed to discriminate global semantic information among different samples, which might not be as crucial for tasks requiring local semantic information such as segmentation. To address this limitation, future research can explore incorporating contrastive learning on local features.
Overall, our algorithm demonstrates superior performance across multiple tasks, including image-sketch, video-audio, and image-depth map segmentation, showcasing its effectiveness and potential for knowledge distillation in cross-modality scenarios.

\begin{table}[H]
    \centering
    \begin{tabular}{c|cc|c|cc}
    \toprule 
        \multirow{2}{*}{Tasks} & \multicolumn{2}{c|}{image-sketch} & video-audio & {image-depth map} \\ \cmidrule{2-5}
         & Sketchy & TUBerlin & VGGSound & NYU-Dpeth V2  \\ \midrule
        TV Distance & 0.04 & 0.04 & 0.10 & 0.06 \\ \midrule
        LE($\Delta$)  & 8.30 & 11.06 & 9.82 & 3.48 \\ 
        FT($\Delta$)  & 1.73 & 3.38 & 2.46 & 3.47 \\
    \bottomrule
    \end{tabular}
    \caption{Comparison of Performance Improvement and Estimated Total Variation Distance. \textbf{LE} means linear evaluation; \textbf{FT} means fine-tuning.}
    \label{tab:d1_dist}
\end{table}

\textbf{Relationship with the generalization bound}. 
Moreover, our experimental results corroborate the bounds derived in Theorem~\ref{thm:whole_alg}. Specifically, we observe that our method exhibits a significant performance improvement through cross-modality distillation in the image-sketch modality, while the improvement in the video-audio modality is more modest. This discrepancy can be attributed to the inherent similarities between images and sketches, which share a common 2-dimensional shape representation. Sketches can be viewed as a direct abstraction of RGB images, leading to a more effective knowledge transfer. On the other hand, video and audio data exhibit larger variations. During training, audio is typically represented by spectrograms, while video is treated as a sequence of images. Consequently, the performance improvement achieved by our method in the video-audio task amounts to less than a 2\% improvement. Conversely, in the image-sketch modalities, our method demonstrates a notable boost in performance, ranging from 2\% to 3\% improvement, even when starting from a higher baseline accuracy. These results just validate the insight in our theoretical results. 
To provide object evidence about our theory, we compute the estimated total variation distances and performance improvements of several datasets and tasks, specifically, we use the performance difference between CMD + LE and SSL + LE as LE($\Delta$), CMD + FT and Sup FT as FT($\Delta$). As shown in Table~\ref{tab:d1_dist}, our discussion in this section actually fits with the estimated TV distance that image-sketch has a smaller modality gap than video-audio. The performance improvements of image-sketch are more significant than video-audio from the results in Table~\ref{tab:d1_dist}. Since the downstream task of image-depthmap datasets is segmentation which is different from the classification task as image-sketch and video-audio, it is unfair to compare the results directly with the other two cases. 

% These results highlight the effectiveness of our approach in leveraging the shared structure and information between images and sketches, leading to enhanced performance in cross-modality distillation tasks.
% Furthermore, our experimental results align well with the bounds derived in Theorem \ref{thm:whole_alg}. Specifically, we observe that our method achieves a greater performance improvement through cross-modality distillation in modalities such as image-sketch, compared to modalities such as  video-audio. This can be attributed to the inherent similarities between images and sketches, as they both share a common 2-dimensional shape representation, and sketches can be viewed as a direct abstraction of RGB images.
% However, in the case of video and audio data, the variations are much larger compared to images and sketches. During training, audio is typically represented by spectrograms, while video can be seen as a sequence of images. As a result, the performance improvement achieved by our method is more modest in the video-audio task, amounting to less than a 2\% improvement.
% Conversely, in the image-sketch modalities, our method demonstrates a notable boost in performance, ranging from 2\% to 3\% improvement, even when starting from a higher baseline accuracy. 
% \textcolor{red}{Relationship with the MFH.}

\subsection{More discussion about the method}

\begin{table} [H]
    \centering
    \begin{tabular}{c|c|c|c}
    \toprule 
        Models  & ResNet50(1x) & ResNet50(2x) & ResNet50(4x) \\ \midrule
        ResNet18 & 67.06 & 67.64  & 67.34 \\ 
        ResNet50 & 68.72 & 68.90 & 69.58 \\
    \bottomrule
    \end{tabular}
    \caption{The results of image-sketch task with different teacher and student networks. We report the top-1 accuracy of downstream classification on the TUBerlin sketch with linear evaluation.}
    \label{tab:dist_model}
    % \vspace{-10pt}
\end{table}
\noindent \textbf{Distillation with more structures}.
Though the modality difference can be seen as the main distillation source in cross-modality distillation, it is still natural to consider distilling a large model into a small model in this case.
Thus, in this part, we mainly study how model sizes or structures affect the performance of our method and illustrate that our method can also be used even when distilling the large teacher to a small student model. In detail, we test our algorithm with ResNet50 (1x/2x/4x) as teacher nets and ResNet18, ResNet50 as the student nets. Here, the ResNet50 (1x/2x/4x) means the original ResNet50 with different widths which is the same setting in \cite{chen2020simple}. From Table~\ref{tab:dist_model}, we can find when the teacher model size becomes larger the distilled student models will have a better performance which just fits with the classical results in knowledge distillation. It shows our algorithm's ability to distill information captured not only by the modality difference but also by the model structure inductive bias. 
% In another way, when the student model size becomes larger the performance still becomes better which just validates our method of distilling the modality-dependent information rather than only benefiting from the distillation with the large model.

\begin{figure}
  \centering
  \begin{subfigure}[b]{0.48\textwidth}
         \centering
         \includegraphics[width=\textwidth]{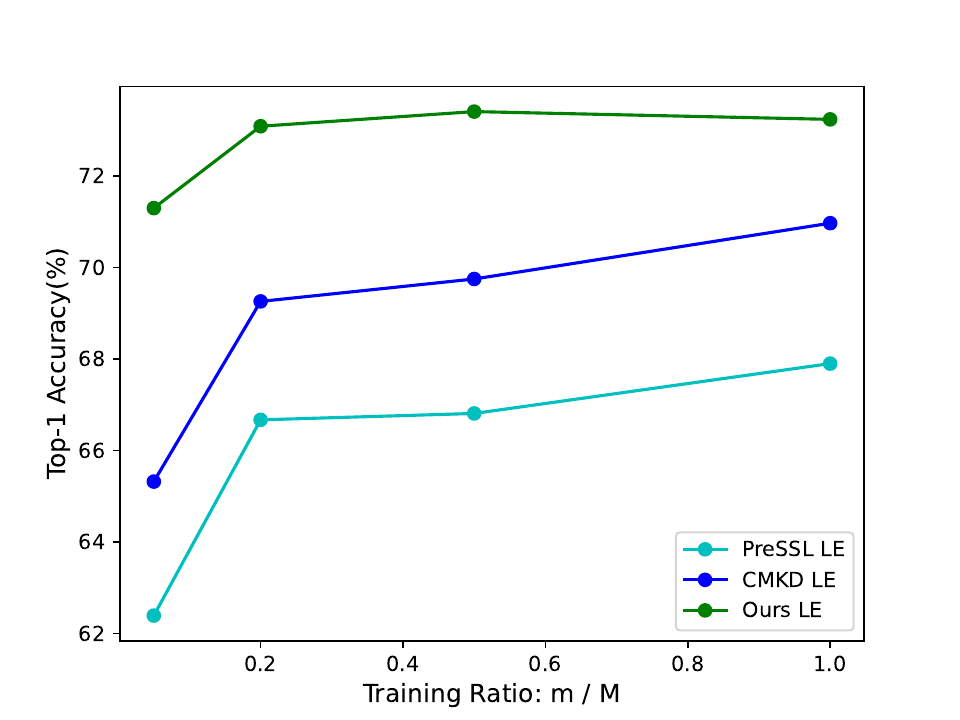}
         \caption{Sketchy}
         \label{fig:m_sketchy}
    \end{subfigure}
    \begin{subfigure}[b]{0.48\textwidth}
         \centering
         \includegraphics[width=\textwidth]{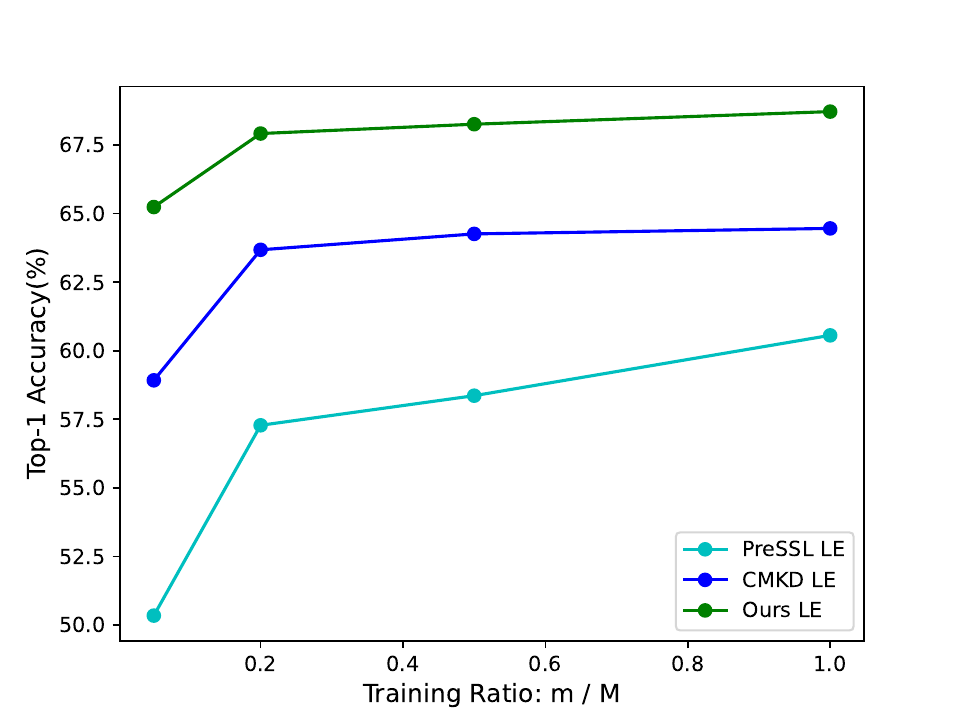}
         \caption{TUBerlin}
         \label{fig:m_tuberlin}
     \end{subfigure}
  \caption{The results of image-sketch tasks with different numbers of distilling samples. The $m / M$ means the percentage of numbers used in distillation. We report the top-1 accuracy of downstream classification on the Sketchy and TUBerlin.}
  \label{fig:distill_ratio}
  % \vspace{-10pt}
\end{figure}
\noindent \textbf{Distillation with less numbers of samples}.
We further test our algorithm with less paired data when cross-modality distilling. We take $m$ as the number of paired data used in distillation and $M$ as the number of whole paired data.  On both Sketchy and TUBerlin, as shown in Figure~\ref{fig:distill_ratio}, our method works well even when the paired data reduces to $m/M=20\%$ of the whole samples. Notably, when we decrease the entire training data to only 5\% of the original training setting, our method experiences a marginal drop of approximately 3\%. In contrast, the performances of CMKD and PreSSL methods decline as the number of training samples decreases, and they exhibit a rapid drop when only 5\% of the data remains. These results highlight the resilience of our algorithm when working with a limited number of paired data for cross-modality distillation.

\begin{table} 
    \centering
    \begin{tabular}{c|c|c|c|c}
    \toprule 
        Models &  $n=1$ & $n=5$ & $n=10$ & Standard(n=60)\\ \midrule
          CMKD LE & 10.21  & 33.18 & 46.27 & 64.46 \\ 
          Ours LE & 19.19 & 42.58 & 52.11 & 68.72\\ \midrule
          Sup FT  & 19.36 & 42.92 & 50.43  &74.48 \\
          CMKD FT & 19.22 & 43.19 & 54.11 & 75.84\\
          Ours FT & 20.29 & 46.53 & 59.90 & 77.44 \\
         
    \bottomrule
    \end{tabular}
    \caption{The results of image-sketch task with different fine-tuning samples, where $n$=1/5/10  means the number of training samples in each class. We report the top-1 accuracy of downstream classification on the TUBerlin sketch. }
    \label{tab:few_shot}
    \vspace{-15pt}
\end{table}
\noindent \textbf{Downstream tasks under few-shot setting}.
To better show that our method can learn meaningful semantic information by cross-modality distillation, we use a few-shot setting test that is used in many self-supervised learning works~\citep{he2020momentum}. Specifically, we test our algorithm on image-sketch modalities and fine-tuning on TUBerlin with each class consisting of only 1,5,10 samples. It can be observed in Table~\ref{tab:few_shot} that when $n$ is small our method even with linear evaluation can achieve comparable performance to the supervised full fine-tuning and our methods just outperforms a lot when we use full fine-tuning as well. This indicates that our method can distill efficient semantic features from the source modality. 

\begin{table}[H]
    \centering
    \begin{tabular}{c|c|ccccc}
    \toprule 
        {Datasets}&{Methods} & \multicolumn{5}{c}{$\alpha$CMC + (1-$\alpha$)CMD} \\
        &&$\alpha$=0&$\alpha$=0.25 &$\alpha$=0.5&$\alpha$=0.75 &$\alpha$=1 \\  \midrule
         \multirow{2}{*}{Sketchy} &{LE}& 72.61 & 72.97 &\textbf{73.45} & 73.19 & 73.24  \\
         &FT& 85.63 & 85.32 & 86.91 & 87.34 & \textbf{87.54}\\ \midrule
         \multirow{2}{*}{TUBerlin} &{LE}& 65.70 & 67.04 & 67.54 & 67.62& \textbf{68.72}\\ 
          &{FT}& \textbf{77.86} & 77.36 & 77.18 & 77.62 &77.44 \\ 
    \bottomrule
    \end{tabular}
    \caption{Comparison of combination of CMC and CMD losses. \textbf{LE} means linear evaluation; \textbf{FT} means fine-tuning.}
    \label{tab:loss_inter}
\end{table}
\noindent \textbf{Interpolation of CMC and CMD losses}. We further conduct an ablation study to investigate how the combination affects the final performance. From Table~\ref{tab:loss_inter}, we can find that combining CMC and CMD losses may help the distillation on some tasks, but there is no dominant choice for every task. Theoretically, I think the combination will make the analysis much more difficult without any more assumptions. We think this is beyond the contribution of the work and can be investigated in future work.

\section{Conclusion}
In this paper, we design a cross-modality contrastive distillation (CMCD) framework for transferring generalizable features from a major source modality to a minor target modality. The proposed framework leverages contrastive learning to fully investigate the positive and negative relationships behind the paired data. Comprehensive experiments covering various modalities (e.g., images, sketches, depth maps, videos, and audio) and tasks (e.g., recognition and segmentation) shows that our algorithms outperform other methods consistently.
Furthermore, we provide a convergence analysis that reveals the test error in the target modality will be bounded by the distance between source and target modalities in our algorithm. These findings underscore the effectiveness and versatility of CMCD as a means of achieving robust feature transfer in various real-world scenarios.

% Comprehensive experiments on various modalities and tasks and theoretical analysis underscore the effectiveness and versatility of CMCD as a means of achieving robust feature transfer in various real-world scenarios.
% Furthermore, we provide a convergence analysis that reveals the test error in the target modality will be bounded by the distance between source and target modalities in our algorithm. These findings 


\bibliography{iclr2024_conference}
\bibliographystyle{iclr2024_conference}

\newpage
\appendix
\section{Detailed Proof of the Lemmas and Theorems}

\subsection{Proof of the Theorem~\ref{lem:con_tv}}
\label{sec:proof_lem_tv}
\begin{proof}
    We just following the proof in the \cite{ge2023provable}.
    First, we reformulate the optimization problem in \eqref{eq:step1_1} as 
    \begin{equation}
        \hat{\phi}_{\mcA} = \argmax_{\phi \in \Phi_{\mcA}} \sum_{i,j=1}^{n_{\mcA}} \log p_{\phi}(\phi(\bm{x}_{ij}^{\mcA}), s_{ij})
    \end{equation}
    % \textcolor{red}{Why the definition of $p_{\phi}$ is reasonable?}
    where $\phi(\bm{x}_{ij}) = (z_i^{\mcA}, z_j^{\mcA}) = (\phi{(x_i^{\mcA})}, \phi{(x_j^{\mcA})})$, and  
  \begin{align}
    p_{\phi}(\phi(\bm{x}_{ij}^{\mcA}), s_{ij}) = \frac{\exp(z_i^{\mcA} \cdot z_j^{\mcA}/\tau)}{\sum_{t} \exp(z_t^{\mcA} \cdot z_j^{\mcA}/\tau)}
  \end{align}
  the Gibbs distribution for the paired data. 
 In fact, it just formulates the cross-modality contrastive learning framework by the maximum likelihood estimation (MLE). We ignore the $\mcA$ subscripts/upscripts and the side information $s$ for notation simplicity in the proof.
    By the definition of $\hat{\phi}$, we have
    \begin{align}
        0  &\le \frac{1}{2} \left(\sum_{i, j=1}^{n} \log p_{\hat{\phi}}(\bm{x}_{ij}) - \sum_{i, j=1}^{n} \log p_{\phi^{*}}(\bm{x}_{ij})\right)\\
            & = \frac{1}{2} \sum_{i, j=1}^{n} \log \frac{p_{\hat{\phi}}(\bm{x}_{ij})}{p_{\phi^{*}}(\bm{x}_{ij})}\\
    \end{align}
    To construct the relationship between $d_{TV}$ and the previous formula, we use Markov inequality and Boole inequality (subadditivity of events). Recall that we define $\mathcal{P}_{\mathcal{X} \times \mathcal{S}}(\Phi) = \{p_{\phi}(\bm{x}, s)|\phi \in \Phi\}$ as the possible distribution family of $\Phi$. For notation simplicity, we denote $\mathcal{P}_{\mathcal{X}_{\mcA} \times \mathcal{S}}(\Phi_{\mcA})$ as $\mathcal{P}$ in this proof. Then we denote the $\epsilon$-bracket class as $\mathcal{N}_{[]}(\mathcal{P}, \epsilon)$, $N_{[]}(\mathcal{P}, \epsilon) = |\mathcal{N}_{[]}(\mathcal{P}, \epsilon)|$.
    % (Ok, pay attention to the $N$ and $\mathcal{N}$). 
    For any $\overline{p}_{\phi} \in \mathcal{N}_{[]}(\mathcal{P}, \epsilon)$, we have the following Markov inequality,
    \begin{align}
        \mathbb{P}(\exp({\frac{1}{2} \sum_{i,j=1}^{n} \log \frac{\overline{p}_{{\phi}}(\bm{x}_{ij})}{p_{\phi^{*}}(\bm{x}_{ij})}} \ge t)) \le \frac{\mathbb{E}[\exp({\frac{1}{2} \sum_{i,j=1}^{n} \log \frac{\overline{p}_{\hat{\phi}}(\bm{x}_{ij})}{p_{\phi^{*}}(\bm{x}_{ij})}})]}{t}\\
        \mathbb{P}\left(\exp\left({\frac{1}{2} \sum_{i,j=1}^{n} \log \frac{\overline{p}_{{\phi}}(\bm{x}_{ij})}{p_{\phi^{*}}(\bm{x}_{ij})}} \ge \frac{C\mathbb{E}[\exp({\frac{1}{2} \sum_{i,j=1}^{n} \log \frac{\overline{p}_{\hat{\phi}}(\bm{x}_{ij})}{p_{\phi^{*}}(\bm{x}_{ij})}})]}{\delta}\right)\right) \le \delta / C\\
        % \mathbb{P}(e^{\frac{1}{2} \sum_{i,j=1}^{n} \log \frac{\overline{p}_{\hat{\phi}}(\bm{x}_{ij})}{p_{\phi^{*}}(\bm{x}_{ij})}} \le \frac{C\mathbb{E}[e^{\frac{1}{2} \sum_{i,j=1}^{n} \log \frac{\overline{p}_{\hat{\phi}}(\bm{x}_{ij})}{p_{\phi^{*}}(\bm{x}_{ij})}}]}{\delta} ) \ge 1 - \delta / C
    \end{align}
    Define the event $D_{\overline{p}_{\phi}}$ as 
    \begin{equation}
        D_{\overline{p}_{\phi}} = \{\bm{x}: \exp\left({\frac{1}{2} \sum_{i,j=1}^{n} \log \frac{\overline{p}_{{\phi}}(\bm{x}_{ij})}{p_{\phi^{*}}(\bm{x}_{ij})}}\right) \ge \frac{C\mathbb{E}[\exp({\frac{1}{2} \sum_{i,j=1}^{n} \log \frac{\overline{p}_{{\phi}}(\bm{x}_{ij})}{p_{\phi^{*}}(\bm{x}_{ij})}})]}{\delta}\}
    \end{equation}
    Then by iterating over all $\overline{p}_{\phi} \in \mathcal{N}_{[]}(\mathcal{P}_{A}, \epsilon)$ we have, 
    \begin{align}
        \mathbb{P}(\cup_{\overline{p}_{\phi} \in \mathcal{N}_{[]}(\mathcal{P}_{A}, \epsilon)} D_{\overline{p}_{\phi}}) &\le \sum_{\overline{p}_{\phi} \in \mathcal{N}_{[]}(\mathcal{P}_{A}, \epsilon)} \mathbb{P}(D_{\overline{p}_{\phi}})\\
        & \le \frac{N_{[]}(\mathcal{P}_{A}, \epsilon) \cdot \delta}{C}
    \end{align}
    Take $C = N_{[]}(\mathcal{P}, \epsilon)$, we have with probability at least $1 - \delta$, for all $\overline{p}_{\phi} \in \mathcal{N}_{[]}(\mathcal{P}, \epsilon)$
    \begin{align}
        \exp\left({\frac{1}{2} \sum_{i,j=1}^{n} \log \frac{\overline{p}_{{\phi}}(\bm{x}_{ij})}{p_{\phi^{*}}(\bm{x}_{ij})}}\right) \le \mathbb{E} [\exp({\frac{1}{2} \sum_{i,j=1}^{n} \log \frac{\overline{p}_{{\phi}}(\bm{x}_{ij})}{p_{\phi^{*}}(\bm{x}_{ij})}})] \cdot \frac{N_{[]}(\mathcal{P}, \epsilon)}{\delta}\\
        {\frac{1}{2} \sum_{i,j=1}^{n} \log \frac{\overline{p}_{{\phi}}(\bm{x}_{ij})}{p_{\phi^{*}}(\bm{x}_{ij})}} \le \log \mathbb{E} [\exp({\frac{1}{2} \sum_{i,j=1}^{n} \log \frac{\overline{p}_{{\phi}}(\bm{x}_{ij})}{p_{\phi^{*}}(\bm{x}_{ij})}})] + \log \frac{N_{[]}(\mathcal{P}, \epsilon)}{\delta}
    \end{align}
    By the definition of bracket class, $\overline{p}_{\hat{\phi}}$ satisfies, with probability at least $1-\delta$
    \begin{align}
        0 \le {\frac{1}{2} \sum_{i,j=1}^{n} \log \frac{\overline{p}_{\hat{\phi}}(\bm{x}_{ij})}{p_{\phi^{*}}(\bm{x}_{ij})}} &\le \log \mathbb{E} [\exp({\frac{1}{2} \sum_{i,j=1}^{n} \log \frac{\overline{p}_{\hat{\phi}}(\bm{x}_{ij})}{p_{\phi^{*}}(\bm{x}_{ij})}})] + \log \frac{N_{[]}(\mathcal{P}, \epsilon)}{\delta}\\
        &= \sum_{i,j=1}^{n} \log \mathbb{E}[\sqrt{\frac{\overline{p}_{\hat{\phi}}(\bm{x}_{ij})}{p_{\phi^{*}}(\bm{x}_{ij})}}] + \log \frac{N_{[]}(\mathcal{P}, \epsilon)}{\delta} \\
        &= m^2 \log \int\sqrt{{\overline{p}_{\hat{\phi}}(\bm{x})}\cdot{p_{\phi^{*}}(\bm{x})}}d\bm{x} + \log \frac{N_{[]}(\mathcal{P}, \epsilon)}{\delta} \\
        &\le m^2 (\int \sqrt{{\overline{p}_{\hat{\phi}}(\bm{x})}\cdot{p_{\phi^{*}}(\bm{x})}}d\bm{x} - 1) + \log \frac{N_{[]}(\mathcal{P}, \epsilon)}{\delta} \\
    \end{align}
    By rearranging the terms, 
    \begin{align}
        1 - \int \sqrt{{\overline{p}_{\hat{\phi}}(\bm{x})}\cdot{p_{\phi^{*}}(\bm{x})}}d\bm{x} \le \frac{1}{m^2}\log \frac{N_{[]}(\mathcal{P}_{A}, \epsilon)}{\delta} \\ 
        \int \left(\sqrt{{\overline{p}_{\hat{\phi}}(\bm{x})}} - \sqrt{{p_{\phi^{*}}(\bm{x})}}\right)^2d\bm{x} \le \frac{2}{m^2}\log \frac{N_{[]}(\mathcal{P}_{A}, \epsilon)}{\delta} \\ 
    \end{align}
    By the definition of $\epsilon$-bracket class, we have
    \begin{align}
        \int \left(\sqrt{{\overline{p}_{\hat{\phi}}(\bm{x})}} + \sqrt{{p_{\phi^{*}}(\bm{x})}}\right)^2dx & \le 2 + 2\int \sqrt{{\overline{p}_{\hat{\phi}}(\bm{x})}\cdot{p_{\phi^{*}}(\bm{x})}} dx \\
        & \le 2 + \int \overline{p}_{\hat{\phi}}(\bm{x}) + p_{\phi^{*}}(\bm{x}) dx \\ 
        & \le 2 + 2(\epsilon + 1) = 2\epsilon + 4 
    \end{align}
    Now we can bound the $d_{TV}$ by Cauchy-Schwarz inequality, with probability at least $1 - \delta$ 
    \begin{align}
        d_{TV}\left(\mathbb{P}_{\hat{\phi}}(\bm{x}), \mathbb{P}_{\phi^{*}}(\bm{x})\right) & = \frac{1}{2} \int |p_{\hat{\phi}}(\bm{x}) - p_{{\phi^{*}}}(\bm{x})| d\bm{x}\\
        &\le\frac{1}{2}  \int |\overline{p}_{\hat{\phi}}(\bm{x}) - p_{{\phi^{*}}}(\bm{x})| d\bm{x} + \frac{1}{2} \int |p_{\hat{\phi}}(\bm{x}) - \overline{p}_{\hat{\phi}}(\bm{x})| d\bm{x}\\
        &\le \frac{1}{2} \left(\int \left(\sqrt{{\overline{p}_{\hat{\phi}}(\bm{x})}} - \sqrt{{p_{\phi^{*}}(\bm{x})}}\right)^2d\bm{x} \cdot \int \left(\sqrt{{\overline{p}_{\hat{\phi}}(\bm{x})}} + \sqrt{{p_{\phi^{*}}(\bm{x})}}\right)^2d\bm{x}\right)^{1/2} + \frac{\epsilon}{2}   \\
        &\le \frac{1}{2}\sqrt{\frac{2}{n^2}\log \frac{N_{[]}(\mathcal{P}, \epsilon)}{\delta} \cdot (2\epsilon + 4) } +  \frac{\epsilon}{2}
    \end{align}
    set $\epsilon = \frac{1}{m^2}$ we can bound the formula above by 
    \begin{equation}
         d_{TV}\left(\mathbb{P}_{\hat{\phi}}(\bm{x}), \mathbb{P}_{\phi^{*}}(\bm{x})\right) \le 3  \sqrt{\frac{1}{n^2}\log \frac{N_{[]}(\mathcal{P}, \frac{1}{n^2})}{\delta}} 
    \end{equation}
 
\end{proof}
\subsection{Proof of the Theorem~\ref{thm:dist}}
\label{sec:proof_con_dist}
\begin{proof}
    The proof is mainly from Chap. 6 in \cite{zhang_2023}.\\ 
    First, we define
    \begin{align}
     \epsilon(\mathcal{L} \circ {\Phi_{\mcB}}, S_m^2) = \sup_{\phi_{\mcB} \in \Phi_{\mcB}} [\mathbb{E}[\mathcal{L}(\hat{\phi}_{\mcA}, {\phi}_{\mcB}, \bm{x}, s)] - \frac{1}{m^2}\sum_{i,j=1}^{m}\mathcal{L}(\hat{\phi}_{\mcA}, {\phi}_{\mcB}, \bm{x}_{ij}, s_{ij})]   
    \end{align}
    and 
    \begin{align}
        \epsilon_n(\mathcal{L} \circ {\Phi_{\mcB}}) = \mathbb{E}_{S_m^2} \epsilon(\mathcal{L} \circ {\Phi_{\mcB}}, S_m^2)   
    \end{align}
     where $S_m^2 = \{(x_i^{\mcA}, x_i^{\mcB})\}_m \times \{(x_i^{\mcA}, x_i^{\mcB})\}_m$. Then by the symmetrization, we can prove 
    \begin{equation}
        \epsilon_n(\mathcal{L} \circ {\Phi_{\mcB}}) \le 2 R_n(\mathcal{L} \circ {\Phi_\mcB})
        \label{eq:d_r}
    \end{equation}
    we do not give detailed proof, readers can refer to Theorem 6.3 in \cite{zhang_2023}.
    Consider $f(\bm{x}_{11}, \dots, \bm{x}_{mm}) = \epsilon(\mathcal{L} \circ {\Phi_{\mcB}}, S_m^2)$, 
    by the assumption that \\ $\sup_{\phi_{\mcB} \in \Phi_\mcB, \bm{x}_{ij}} \langle {\phi_{\mcB}}(\bm{x}_i), {\phi_{\mcB}}(\bm{x}_j)\rangle  \le B$, we can check the condition for McDiarmid's inequality,
    \begin{align}
        \sup_{\bm{x}_{11}, \dots, \bm{x}_{mm}, \bm{x}_{ij}^{\prime}} |f(\bm{x}_{11}, \dots, \bm{x}_{ij}^{\prime},&\dots, \bm{x}_{mm}) - f(\bm{x}_{11}, \dots, \bm{x}_{ij}^{\prime}, \dots, \bm{x}_{mm})| \\
        &\le \frac{1}{m^2} \sup_{\bm{x}_{ij}, \bm{x}_{ij}^{\prime}} |\sup_{\phi_{\mcB}}[\mathcal{L}(\hat{\phi}_{\mcA}, {\phi}_{\mcB}, \bm{x}_{ij}^{\prime}, s_{ij}^{\prime}) - \mathcal{L}(\hat{\phi}_{\mcA}, {\phi}_{\mcB}, \bm{x}_{ij}, s_{ij})]|\\
        &\le \frac{1}{m^2}\sup_{\bm{x}_{ij}, \bm{x}_{ij}^{\prime}} \sup_{\phi_{\mcB}}|\mathcal{L}(\hat{\phi}_{\mcA}, {\phi}_{\mcB}, \bm{x}_{ij}^{\prime}, s_{ij}^{\prime}) - \mathcal{L}(\hat{\phi}_{\mcA}, {\phi}_{\mcB}, \bm{x}_{ij}, s_{ij})|\\
        &\le \frac{1}{m^2} 2|\log \frac{1+\exp(B)}{1+\exp(-B)}| \label{eq:B_trick}\\
        &= \frac{2B}{m^2}\\
    \end{align}
    then we can apply McDiarmid's inequality, 
    \begin{align}
        % \mathbb{P}(f(x_1, \dots, x_n) \le \mathbb{E}_{S_n}f(x_1, \dots, x_n) - \epsilon) \le \exp (\frac{-2n\epsilon^2}{4L^2})\\
        \mathbb{P}(f(\bm{x}_{11}, \dots, \bm{x}_{mm}) \ge \mathbb{E}_{S_m^2}f(\bm{x}_{11}, \dots, \bm{x}_{mm}) + \epsilon) \le \exp (\frac{-m^2\epsilon^2}{2B^2})
    \end{align}
    then with probability at least $1 - \delta$, 
    \begin{align}
        f(\bm{x}_{11}, \dots, \bm{x}_{mm}) \le \mathbb{E}_{S_m^2}f(\bm{x}_{11}, \dots, \bm{x}_{mm}) + B\sqrt{\frac{2\ln (1/ \delta)}{m^2}}\\
        \sup_{\phi_{\mcB} \in \Phi_{\mcB}} [\mathbb{E}[\mathcal{L}(\hat{\phi}_{\mcA}, {\phi}_{\mcB}, \bm{x}, s)] - \frac{1}{m^2}\sum_{i,j=1}^{m}\mathcal{L}(\hat{\phi}_{\mcA}, {\phi}_{\mcB}, \bm{x}_{ij}, s_{ij})]    \le \epsilon_n(\mathcal{L} \circ {\Phi_{\mcB}}) + B\sqrt{\frac{2\ln (1/ \delta)}{m^2}}
    \end{align}
    Combined with the result of \eqref{eq:d_r}, we have with probability at least $1 - \delta$, for any $\phi_{\mcB} \in \Phi_{\mcB}$,
    \begin{equation}
        \mathbb{E}[\mathcal{L}(\hat{\phi}_{\mcA}, \hat{\phi}_{\mcB}, \bm{x}, s)] - \frac{1}{m^2}\sum_{i,j=1}^{m}\mathcal{L}(\hat{\phi}_{\mcA}, \hat{\phi}_{\mcB}, \bm{x}_{ij}, s_{ij}) \le 2 R_n(\mathcal{L} \circ {\Phi_{\mcB}}) + B\sqrt{\frac{2\ln (1/ \delta)}{m^2}}
    \end{equation}
    A similar discussion shows that with probability at least $1 - \delta$, $\phi_{\mcB} \in \Phi_{\mcB}$,
    \begin{equation}
           \frac{1}{m^2}\sum_{i,j=1}^{m}\mathcal{L}(\hat{\phi}_{\mcA}, \hat{\phi}_{\mcB}, \bm{x}_{ij}, s_{ij}) - \mathbb{E}[\mathcal{L}(\hat{\phi}_{\mcA}, \hat{\phi}_{\mcB}, \bm{x}, s)]  \le 2 R_n(\mathcal{L} \circ {\Phi_{\mcB}}) + B\sqrt{\frac{2\ln (1/ \delta)}{m^2}}
    \end{equation}
\end{proof}

\subsection{Proof of the Theorem~\ref{thm:whole_alg}}
\label{sec:proof_whole}

We first introduce a lemma.
\begin{lemma}[Bound of ERM.]
    \label{lem:erm}
     Suppose that $\mathcal{L}(\cdot, \cdot)$ is a $L$-bounded loss function. Given a fixed $\phi \in \Phi$, with probability at least $1-\delta$, for any $\psi \in \Psi$,
     \begin{equation}
        \mathbb{E}[\mathcal{L}({\psi \circ \phi}(x), y)] - \frac{1}{n}\sum_{i=1}^{n}\mathcal{L}({\psi \circ \phi}(x), y) \le 2 R_n(\mathcal{L} \circ { \Psi \circ \phi}) + L\sqrt{\frac{2\ln (1/ \delta)}{n}}
    \end{equation}
    % and
    % \begin{equation}
    %     |\mathbb{E}[\mathcal{L}({\psi \circ \phi}(x), y)] - \frac{1}{n}\sum_{i=1}^{n}\mathcal{L}({\psi \circ \phi}(x), y)| \le 2 R_n(l \circ {\psi \circ \Phi}) + L\sqrt{\frac{2\ln (2/ \delta)}{n}}
    % \end{equation}
    % where $\mathbb{E}:=\mathbb{E}_{\mathbb{P}_{{\phi}_{\mcB}^{*}, {\psi}_{\mcB}^{*}}}$
\end{lemma}
\begin{proof}
    This proof is almost the same as the proof in Theorem~\ref{thm:dist}.
    First, we define $\epsilon(\mathcal{L} \circ {\Psi \circ \phi}, S_n) = \sup_{\psi \in \Psi} [\mathbb{E}[\mathcal{L}({\psi \circ \phi}(x), y)] - \frac{1}{n}\sum_{i=1}^{n}\mathcal{L}({ \psi \circ \phi}(x), y)]$ and $\epsilon_n(\mathcal{L} \circ {\Psi \circ \phi}) = \mathbb{E}_{S_n} \epsilon(\mathcal{L} \circ {\Psi \circ \phi}, S_n)$ where $S_n = \{(x_i, y_i)\}_n$. Then by the symmetrization, we can prove 
    \begin{equation}
        \epsilon_n(\mathcal{L} \circ {\Psi \circ \phi}) \le 2 R_n(\mathcal{L} \circ {\Psi \circ \phi})
        \label{eq:2d_r}
    \end{equation}
    we do not give detailed proof, readers can refer to Theorem 6.3 in \cite{zhang_2023}.
    Consider $f(X_1, \dots, X_n) = \sup_{\psi \in \Psi} [\mathbb{E}[\mathcal{L}({\psi \circ \phi}(x), y)] - \frac{1}{n}\sum_{i=1}^{n}\mathcal{L}({\psi \circ \phi}(x), y)]$, it is obvious that $\sup_{x_1, \dots, x_n, x_i^{\prime}} |f(x_1, \dots, x_i, \dots, x_n) - f(x_1, \dots, x_i^{\prime}, \dots, x_n)| \le \frac{2}{n}L$, then we can apply McDiarmid's inequality, 
    \begin{align}
        % \mathbb{P}(f(x_1, \dots, x_n) \le \mathbb{E}_{S_n}f(x_1, \dots, x_n) - \epsilon) \le \exp (\frac{-2n\epsilon^2}{4L^2})\\
        \mathbb{P}(f(x_1, \dots, x_n) \ge \mathbb{E}_{S_n}f(x_1, \dots, x_n) + \epsilon) \le \exp (\frac{-n\epsilon^2}{2L^2})
    \end{align}
    then with probability at least $1 - \delta$, 
    \begin{align}
        f(x_1, \dots, x_n) \le \mathbb{E}_{S_n}f(x_1, \dots, x_n) + L\sqrt{\frac{2\ln (1/ \delta)}{n}}\\
        \sup_{\psi \in \Psi} [\mathbb{E}[\mathcal{L}({\psi \circ \phi}(x), y)] - \frac{1}{n}\sum_{i=1}^{n}\mathcal{L}({\psi \circ \phi}(x), y)] \le \epsilon_n(\mathcal{L} \circ {\Psi \circ \phi}) + L\sqrt{\frac{2\ln (1/ \delta)}{n}}
    \end{align}
    Combined with the result of \eqref{eq:d_r}, we have with probability at least $1 - \delta$, for any $\psi \in \Psi$,
    \begin{equation}
        \mathbb{E}[\mathcal{L}({\psi \circ \phi}(x), y)] - \frac{1}{n}\sum_{i=1}^{n}\mathcal{L}({\psi \circ \phi}(x), y) \le 2 R_n(\mathcal{L} \circ {\Psi \circ \phi}) + L\sqrt{\frac{2\ln (1/ \delta)}{n}}
    \end{equation}
    A similar discussion shows that with probability at least $1 - \delta$, for any $\psi \in \Psi$,
    \begin{equation}
        \frac{1}{n}\sum_{i=1}^{n}\mathcal{L}({\psi \circ \phi}(x), y) - \mathbb{E}[\mathcal{L}({\psi \circ \phi}(x), y)] \le 2 R_n(\mathcal{L} \circ {\Psi \circ \phi}) + L\sqrt{\frac{2\ln (1/ \delta)}{n}}
    \end{equation}
    take $1 - \delta / 2$ for each inequality and combine the results, we get with probability at least $1 - \delta$, for any $\psi \in \Psi$,
    \begin{equation}
        |\mathbb{E}[\mathcal{L}({\psi \circ \phi}(x), y)] - \frac{1}{n}\sum_{i=1}^{n}\mathcal{L}({\psi \circ \phi}(x), y)| \le 2 R_n(\mathcal{L} \circ {\Psi \circ \phi}) + L\sqrt{\frac{2\ln (2/ \delta)}{n}}
    \end{equation}
\end{proof}

Now we prove the Theorem~\ref{thm:whole_alg},
\begin{proof}
    The proof starts from the standard convergence analysis with Rademacher complexity. 
    By the Lemma \ref{lem:erm}, given the fixed ${\hat{\phi}_\mcB}(x)$, we have with probability at least $1 - \delta$,
    \begin{equation}
        \label{eq:uni_psi:A}
        \mathbb{E}[\mathcal{L}({\hat{\psi}_{\mcB} \circ \hat{\phi}_{\mcB}}(x), y)] \le \frac{1}{n} \sum_{i=1}^{n_{\mcB}} \mathcal{L}({\hat{\psi}_{\mcB} \circ \hat{\phi}_{\mcB}}(x), y) + 2 R_n(\mathcal{L} \circ {\Psi_{\mcB} \circ \hat{\phi}_{\mcB}}) + L\sqrt{\frac{2\ln (1/ \delta)}{n_{\mcB}}}
    \end{equation}
    we only need to handle the empirical risk $ \frac{1}{n} \sum_{i=1}^{n_{\mcB}} \mathcal{L}({\hat{\phi}_{\mcB}, \hat{\psi}_{\mcB}}(x), y)$,  
    % here we use $\mathbb{E}:=\mathbb{E}_{\mathbb{P}_{{\phi}_{\mcB}^{*}, {\psi}_{\mcB}^{*}}}$ for notation simplicity,
    from the definition of $\hat{\psi}_{\mcB}$ in \eqref{eq:step3} we get,
    \begin{align}
       \frac{1}{n} \sum_{i=1}^{n_{\mcB}} \mathcal{L}({\hat{\psi}_{\mcB} \circ \hat{\phi}_{\mcB}}(x), y) &\le \epsilon_{\mcB} + \frac{1}{n} \sum_{i=1}^{n_{\mcB}} \mathcal{L}({ {\psi}_{\mcB}^{*} \circ \hat{\phi}_{\mcB}}(x), y) - \mathbb{E}[\mathcal{L}({{\psi}_{\mcB}^{*} \circ \hat{\phi}_{\mcB}}(x), y)]  \label{eq:con:A}\\
       &+ \mathbb{E}[\mathcal{L}({{\psi}_{\mcB}^{*} \circ \hat{\phi}_{\mcB}}(x), y)]\label{eq:lat:A}
       % - \mathbb{E}[\mathcal{L}({{\phi}_{\mcB}^{*}, {\psi}_{\mcB}^{*}}(x), y)] \label{eq:lat_sA}\\
       % &+ \mathbb{E}[\mathcal{L}({{\phi}_{\mcB}^{*}, {\psi}_{\mcB}^{*}}(x), y)] \label{eq:opt_sA}
    \end{align}
    the first term~\eqref{eq:con:A} can be bounded by concentration inequality and we only need to bound the second term~\eqref{eq:lat:A} further.
    By the assumption \ref{ass1},
    \begin{align}
        \mathbb{E}[\mathcal{L}({\hat{\phi}_{\mcB}, {\psi}_{\mcB}^{*}}(x), y)] &\le \kappa\mathbb{E}[\mathbb{E}_{x^{\prime}}[\mathcal{L}_{\text{CMD}}(\hat{\phi}_{\mcB}, {\phi}_{\mcB}^{*}, (x, x^{\prime}), s)]] \label{eq:E_start} \\
        &= \kappa \mathbb{E}[\mathcal{L}_{\text{CMD}}(\hat{\phi}_{\mcB}, {\phi}_{\mcB}^{*}, (x, x^{\prime}), s)] \quad (\text{denote } \bm{x}=(x, x^{\prime})) \\ 
        &= \kappa (\mathbb{E}[\mathcal{L}_{\text{CMD}}(\hat{\phi}_{\mcB}, {\phi}_{\mcB}^{*}, \bm{x}, s)] - \mathbb{E}[\mathcal{L}_{\text{CMD}}(\hat{\phi}_{\mcB}, {\phi}_{\mcA}^{*}, \bm{x}, s)] \\
        &\quad + \mathbb{E}[\mathcal{L}_{\text{CMD}}(\hat{\phi}_{\mcB}, {\phi}_{\mcA}^{*}, \bm{x}, s)] - \mathbb{E}[\mathcal{L}_{\text{CMD}}(\hat{\phi}_{\mcB}, \hat{\phi}_{\mcA}, \bm{x}, s)] \\
        &\quad+ \mathbb{E}[\mathcal{L}_{\text{CMD}}(\hat{\phi}_{\mcB}, \hat{\phi}_{\mcA}, \bm{x}, s)] ) \\
        &= \kappa(\mathbb{E} [-p_{{\phi}_{\mcB}^{*}}(\bm{x}, s)\log p_{\hat{\phi}_{\mcB}}(\bm{x}, s)+ p_{{\phi}_{\mcA}^{*}}(\bm{x}, s)\log p_{\hat{\phi}_{\mcB}}(\bm{x}, s)]\\
        &\quad + \mathbb{E} [-p_{{\phi}_{\mcA}^{*}}(\bm{x}, s)\log p_{\hat{\phi}_{\mcB}}(\bm{x}, s)+ p_{\hat{\phi}_{\mcA}}(\bm{x}, s)\log p_{\hat{\phi}_{\mcB}}(\bm{x}, s)]\\
        &\quad + \mathbb{E}[\mathcal{L}_{\text{CMD}}(\hat{\phi}_{\mcB}, \hat{\phi}_{\mcA}, \bm{x}, s)]) \\
        &\le \kappa B \cdot \left(d_{TV}(\mathbb{P}_{{\phi}_{\mcB}^{*}}, \mathbb{P}_{{\phi}_{\mcA}^{*}}) + d_{TV}(\mathbb{P}_{\hat{\phi}_{\mcA}}, \mathbb{P}_{{\phi}_{\mcA}^{*}}) \right) \label{ineq:tv} \\ 
        &\quad + \kappa\mathbb{E}[\mathcal{L}_{\text{CMD}}(\hat{\phi}_{\mcB}, \hat{\phi}_{\mcA}, \bm{x}, s)]\label{eq:E_end} \\
    \end{align}
    where the inequality \eqref{ineq:tv} comes from a same argument as \eqref{eq:B_trick}. 
    
    To derive the final result, define two events,
    \begin{equation}
        D_{\mcA} = \left\{S_n^{\mcA}: d_{TV}(\mathbb{P}_{\hat{\phi}_{\mcA}}(\bm{x}, s), \mathbb{P}_{{\phi}^{*}_{\mcA}}(\bm{x}, s)) \le 3 \sqrt{\frac{1}{{n_{\mcA}}^2} \log \frac{\bracnum(\mathcal{P}_{\mathcal{X}_{\mcA}  \times \mathcal{S}}(\Phi_{\mcA}), \frac{1}{{n_{\mcA}}^2})}{\delta}}\right\}
    \end{equation}

    \begin{equation}
        D_{\mcA\mcB} = \left\{S_m^2: \mathbb{E}[\mathcal{L}(\hat{\phi}_{\mcB}, \hat{\phi}_{\mcA}, \bm{x}, s)] - \frac{1}{m^2}\sum_{i,j=1}^{m}\mathcal{L}_{\text{CMD}}(\hat{\phi}_{\mcB}, \hat{\phi}_{\mcA}, \bm{x}_{ij},s_{ij}) \le 2 R_n(\mathcal{L}_{\text{CMD}} \circ {\Phi_{\mcB}}) + L\sqrt{\frac{2\ln (1/ \delta)}{m^2}}\right\}
    \end{equation}
    By the Theorem \ref{thm:dist} and Lemma \ref{lem:con_tv}, we have $\mathbb{P}(D_{\mcA}) \ge 1 - \delta, \mathbb{P}(D_{\mcA\mcB}|\hat{\phi}_{\mcA}) \ge 1 - \delta$, then consider the $\mathbb{P}(D_{\mcA} \cap D_{\mcA\mcB})$,
    \begin{align}
        \mathbb{P}(D_{\mcA} \cap D_{\mcA\mcB}) &= \mathbb{E}[\mathbbm{1}_{D_{\mcA}}\mathbb{P}(D_{\mcA\mcB}|\hat{\phi}_{\mcA})]\\
        &\ge (1-\delta)\cdot \mathbb{P}(D_{\mcA}) \\
        &\ge (1-\delta)^2 \ge 1 - 2 \delta
    \end{align}
    So with probability at least $1-\delta$, 
    \begin{align}
        \mathbb{E}[\mathcal{L}({\hat{\phi}_{\mcB}, {\psi}_{\mcB}^{*}}(x), y)] &\le \kappa B \cdot d_{TV}(p_{{\phi}_{\mcB}^{*}}, p_{{\phi}_{\mcA}^{*}}) \\
        &\quad + 3\kappa B \cdot  \sqrt{\frac{1}{{n_{\mcA}}^2} \ln \frac{2\bracnum(\mathcal{P}_{\mathcal{X}_{\mcA}  \times \mathcal{S}}(\Phi_{\mcA}), \frac{1}{{n_{\mcA}}^2})}{\delta}} \\
        &\quad + \kappa(\epsilon_{\mcA\mcB} + 2 R_{m^2}(\mathcal{L}_{\text{CMD}} \circ {\Phi_{\mcB}}) + L\sqrt{\frac{2\ln (2/ \delta)}{m^2}})\label{eq:phi:AB} 
    \end{align}

    By Chernoff bound, with probability at least $1-\delta$, we have
    \begin{equation}
        \label{eq:che}
        \frac{1}{n_{\mcB}} \sum_{i=1}^{n_{\mcB}} \mathcal{L}({\hat{\phi}_{\mcB}, {\psi}_{\mcB}^{*}}(x), y) - \mathbb{E}[\mathcal{L}({\hat{\phi}_{\mcB}, {\psi}_{\mcB}^{*}}(x), y)] \le L\sqrt{\frac{2\ln (1/ \delta)}{n_{\mcB}}}
    \end{equation}
    Take $1-\delta / 2$ for \eqref{eq:uni_psi:A} and \eqref{eq:che}, we get with probability $1-\delta$,
    \begin{equation}
        \label{eq:uni_phi:A}
        \mathbb{E}[\mathcal{L}({\hat{\phi}_{\mcB}, \hat{\psi}_{\mcB}}(x), y)] \le \epsilon_{\mcB} + \mathbb{E}[\mathcal{L}({\hat{\phi}_{\mcB}, {\psi}_{\mcB}^{*}}(x), y)]  + 2 R_{n_{\mcB}}(\mathcal{L} \circ {\Psi \circ \hat{\phi}_{\mcB}}) + 2L\sqrt{\frac{2\ln (2/ \delta)}{n_{\mcB}}}
    \end{equation}
    Take $1-\delta / 2$ for \eqref{eq:uni_phi:A} and \eqref{eq:phi:AB}, we get with probability $1-\delta$,
    \begin{align}
        \mathbb{E}[\mathcal{L}({\hat{\psi}_{\mcB}} \circ \hat{\phi}_{\mcB}(x)&, y)] \le \kappa B \cdot d_{TV}(\mathbb{P}_{{\phi}_{\mcB}^{*}}, \mathbb{P}_{{\phi}_{\mcA}^{*}}) + \kappa\epsilon_{\mcA\mcB} + \epsilon_{\mcB}\\
        &\quad  + 2 \kappa R_{m^2}(\mathcal{L}_{\text{CMD}} \circ {\Phi_{\mcB}})   + 2 R_{n_{\mcB}}(\mathcal{L} \circ \Psi_{\mcB} \circ {\hat{\phi}_{\mcB}})\\ 
        &\quad + 3\kappa B \cdot  \sqrt{\frac{1}{{n_{\mcA}}^2} \ln \frac{4\bracnum(\mathcal{P}_{\mathcal{X}_{\mcA}  \times \mathcal{S}}(\Phi_{\mcA}), \frac{1}{{n_{\mcA}}^2})}{\delta}} + \kappa L\sqrt{\frac{2\ln (4 / \delta)}{m^2}} + 2L\sqrt{\frac{2\ln (4 / \delta)}{n_{\mcB}}}
    \end{align}
\end{proof}

\section{Discussion About CMC Loss}
\label{app:cmc}
In order to introduce a similar bound for the CMC loss, we introduce a likelihood bound assumption,
\begin{assumption}
\label{ass:cmc}
    For any fixed $\phi$, we have
    \begin{align}
        -\log \frac{p_{\phi_{\mcA}, \phi_{\mcB}}(\bm{x})}{p_{\hat\phi_{\mcA}, \phi_{\mcB}}(\bm{x})} \le - (p_{{\phi_{\mcA}}}(\bm{x}) - p_{{\hat\phi_{\mcA}}}(\bm{x})) \log p_{\phi_{\mcB}}(\bm{x})
    \end{align}
    where 
  \begin{align}
    p_{\phi_{\mcA}, \phi_{\mcB}}(\bm{x})  = \frac{\exp(z_i^{\mcA} \cdot z_j^{\mcB}/\tau)}{\sum_{t} \exp(z_t^{\mcA} \cdot z_j^{\mcB}/\tau)},\quad p_{\phi_{\mcA}}(\bm{x}) = \frac{\exp(z_i^{\mcA} \cdot z_j^{\mcA}/\tau)}{\sum_{t} \exp(z_t^{\mcA} \cdot z_j^{\mcA}/\tau)}
  \end{align}
\end{assumption}
From the proofs above, we can find that changing the CMD loss to CMC loss does not affect the lemmas and theorems other than the final results Theorem~\ref{thm:whole_alg}. Thus, we just show that with the assumption~\ref{ass:cmc} we can get the same result as the Theorem~\ref{thm:whole_alg} with CMC loss.
\begin{proof}
    Noticing that the main difference for CMD and CMC losses are between \eqref{eq:E_start} and \eqref{eq:E_end}. We only discuss the bounded process here and other derivations should be the same.
      \begin{align}
        \mathbb{E}[\mathcal{L}({\hat{\phi}_{\mcB}, {\psi}_{\mcB}^{*}}(x), y)] &\le \kappa\mathbb{E}[\mathbb{E}_{x^{\prime}}[\mathcal{L}_{\text{CMC}}(\hat{\phi}_{\mcB}, {\phi}_{\mcB}^{*}, (x, x^{\prime}), s)]] \label{eq:E_start} \\
        &= \kappa \mathbb{E}[\mathcal{L}_{\text{CMC}}(\hat{\phi}_{\mcB}, {\phi}_{\mcB}^{*}, (x, x^{\prime}), s)] \quad (\text{denote } \bm{x}=(x, x^{\prime})) \\ 
        &= \kappa (\mathbb{E}[\mathcal{L}_{\text{CMC}}(\hat{\phi}_{\mcB}, {\phi}_{\mcB}^{*}, \bm{x}, s)] - \mathbb{E}[\mathcal{L}_{\text{CMC}}(\hat{\phi}_{\mcB}, {\phi}_{\mcA}^{*}, \bm{x}, s)] \\
        &\quad + \mathbb{E}[\mathcal{L}_{\text{CMC}}(\hat{\phi}_{\mcB}, {\phi}_{\mcA}^{*}, \bm{x}, s)] - \mathbb{E}[\mathcal{L}_{\text{CMC}}(\hat{\phi}_{\mcB}, \hat{\phi}_{\mcA}, \bm{x}, s)] \\
        &\quad+ \mathbb{E}[\mathcal{L}_{\text{CMC}}(\hat{\phi}_{\mcB}, \hat{\phi}_{\mcA}, \bm{x}, s)] ) \\
        &= \kappa(\mathbb{E} [-\log \frac{p_{{\phi}_{\mcB}^{*},\hat{\phi}_{\mcB}}}{p_{{\phi}_{\mcA}^{*},\hat{\phi}_{\mcB}}}(\bm{x}, s)] + \mathbb{E} [-\log \frac{p_{{\phi}_{\mcA}^{*},\hat{\phi}_{\mcB}}}{p_{\hat{\phi}_{\mcA},\hat{\phi}_{\mcB}}}(\bm{x}, s)]\\
        &\quad + \mathbb{E}[\mathcal{L}_{\text{CMC}}(\hat{\phi}_{\mcB}, \hat{\phi}_{\mcA}, \bm{x}, s)]) \\
        &\le \kappa(\mathbb{E} [-p_{{\phi}_{\mcB}^{*}}(\bm{x}, s)\log p_{\hat{\phi}_{\mcB}}(\bm{x}, s)+ p_{{\phi}_{\mcA}^{*}}(\bm{x}, s)\log p_{\hat{\phi}_{\mcB}}(\bm{x}, s)]\\
        &\quad + \mathbb{E} [-p_{{\phi}_{\mcA}^{*}}(\bm{x}, s)\log p_{\hat{\phi}_{\mcB}}(\bm{x}, s)+ p_{\hat{\phi}_{\mcA}}(\bm{x}, s)\log p_{\hat{\phi}_{\mcB}}(\bm{x}, s)]\\
        &\quad + \mathbb{E}[\mathcal{L}_{\text{CMC}}(\hat{\phi}_{\mcB}, \hat{\phi}_{\mcA}, \bm{x}, s)]) \\
        &\le \kappa B \cdot \left(d_{TV}(\mathbb{P}_{{\phi}_{\mcB}^{*}}, \mathbb{P}_{{\phi}_{\mcA}^{*}}) + d_{TV}(\mathbb{P}_{\hat{\phi}_{\mcA}}, \mathbb{P}_{{\phi}_{\mcA}^{*}}) \right) \label{ineq:tv} \\ 
        &\quad + \kappa\mathbb{E}[\mathcal{L}_{\text{CMC}}(\hat{\phi}_{\mcB}, \hat{\phi}_{\mcA}, \bm{x}, s)]\label{eq:E_end} \\
    \end{align}
    Then we get the same convergence bound of CMC loss as the CMD loss as shown in Theorem~\ref{thm:whole_alg}
\end{proof}

\textbf{Further improvement}. The assumption~\ref{ass:cmc} in this paper is not trivial or prior to the analysis, further work to this research can focus on a better proof and result with CMC loss.
\section{Detailed Settings of Experiments}
\label{app:set_exp}
In this section, we give the detailed settings of datasets and training. 
In our experiments, all cross-modality distillation using a self-supervised learned ResNet on ImageNet. As mentioned in the paper, we only used the well-trained model provided by the official SimCLR with different structures of ResNet50, ResNet50(2x), and ResNet50(4x) but not using the ImageNet data in the distillation. To clarify the cross-modality distillation process, we give the dataset used for transferring and the detailed setting of the downstream task of each pair of modalities.

\begin{table}[H]
    \centering
    \begin{tabular}{c|c|c|c}
    \toprule
        Training Dataset & Sketchy & TUBerlin & Sketchy-Eval \\
        \midrule
        Train/Test Split & 48,290  & 15,000/5,000 & 60,335/15,146\\
        Paired data $M$ &  -- & 15,000 & --\\
        \midrule
        Optimizer & Adam & Adam & Adam\\
        Optimizer Hyper-parameter & (0.9,0.999) & (0.9,0.999) & (0.9,0.999) \\
        Learning Rate Schedule & None & Multi-Step(60,70,80) & Multi-Step(60,70,80) \\
        Learning Rate & 1e-3 & 1e-3 & 1e-3\\
        Epoch & 100 & 100 & 100 \\
        Batch Size & 64 & 64 & 64\\
        \bottomrule
    \end{tabular}
    \caption{Details of image-sketch Distillation.}
    \label{tab:llm_details}
\end{table}
Since there are multiple sketches corresponding to one image in the Sketchy dataset, we consider all these pairs as positive pairs resulting in a total of 48290 training data. The train/split for TUBerlin and Sketchy-Eval just follows the typical setting used in \cite{yu2017sketch,lin2020sketch}. Sketches in Sketchy-eval may have been trained without labels in distillation.
 
\begin{table}[H]
    \centering
    \begin{tabular}{c|c|c}
    \toprule
        Training Dataset & NYU-Depth V2 & NYU-Depth V2-Eval( Disjoint ) \\
        \midrule
        Train/Test Split & 795  & 795/654 \\
        Paired data $M$ &  795 & -- \\
        \midrule
        Optimizer & Adam & Adam \\
        Optimizer Hyper-parameter & (0.9,0.999) & (0.9,0.999) \\
        Learning Rate Schedule & None  & Multi-Step(60,70,80) \\
        Learning Rate & 1e-3 & 1e-2 \\
        Epoch & 100 & 100 \\
        Batch Size & 16 & 16 \\
        \bottomrule
    \end{tabular}
    \caption{Details of image-depth map Distillation.}
    \label{tab:llm_details}
\end{table}
For the image-depth map task, we only use the training data in NYU-Depth V2 and also use the labeled version in downstream segmentation.
\begin{table}[H]
    \centering
    \begin{tabular}{c|c|c}
    \toprule
        Training Dataset & VGGSound & VGGSound-Eval (Disjoint) \\
        \midrule
        Train/Test Split & 4,625 & 10,000/10,000 \\
        Paired data $M$ &  4,625 & -- \\
        \midrule
        Optimizer & Adam & Adam \\
        Optimizer Hyper-parameter & (0.9,0.999) & (0.9,0.999) \\
        Learning Rate Schedule & None  & Multi-Step(60,70,80) \\
        Learning Rate & 1e-3 & 1e-2 \\
        Epoch & 100 & 100 \\
        Batch Size & 16 & 16 \\
        \bottomrule
    \end{tabular}
    \caption{Details of video-audio Distillation.}
    \label{tab:llm_details}
\end{table}
In this case, we sample 4625 pairs of video and audio, translating the video into 12 frames and audio into spectrograms. A disjoint 10000 audio-only dataset is sampled to fine-tune downstream event classification where another 10000 are used for testing.

\end{document}